\documentclass[10pt,twocolumn,letterpaper]{article}

\usepackage{cvpr}
\usepackage{times}
\usepackage{epsfig}
\usepackage{graphicx}
\usepackage{amsmath}
\usepackage{amssymb,textcomp}
\usepackage{subfigure}

\usepackage[pagebackref=true,breaklinks=true,letterpaper=true,colorlinks,bookmarks=false]{hyperref}

\cvprfinalcopy 

\ifcvprfinal\fi
\begin{document}

\title{Large-scale Supervised Hierarchical Feature Learning for Face Recognition}
\author{Jianguo Li, Yurong Chen\\ Intel Labs China}

\maketitle

\begin{abstract}
   This paper proposes a novel face recognition algorithm based on large-scale supervised hierarchical feature learning.
   The approach consists of two parts: hierarchical feature learning and large-scale model learning.
   The hierarchical feature learning searches feature in three levels of granularity in a supervised way.
   First, face images are modeled by receptive field theory, and the representation is an image with many channels of Gaussian receptive maps.
   We activate a few most distinguish channels by supervised learning.
   Second, the face image is further represented by patches of picked channels, and we search from the over-complete patch pool
   to activate only those most discriminant patches.
   Third, the feature descriptor of each patch is further projected to lower dimension subspace with discriminant subspace analysis.

   Learned feature of activated patches are concatenated to get a full face representation.
   A linear classifier is learned to separate face pairs from same subjects and different subjects.
   As the number of face pairs are extremely large, we introduce ADMM (alternative direction method of multipliers)
   to train the linear classifier on a computing cluster. Experiments show that more training samples will bring notable accuracy improvement.

   We conduct experiments on FRGC and LFW.  Results show that the proposed approach outperforms existing algorithms notably.
   Besides, the proposed approach is small in memory footprint, and low in computing cost, which makes it suitable for embedded applications.

\end{abstract}


\section{Introduction}
Face recognition has received a great deal of attention from research communities and industries over the past two decades
due to its wide range of applications \cite{StanFace,ZhaoFRSurvey}.
Recently, with the evolution of handheld digital devices and social networks, face recognition enters a new era for applications in handheld devices and social networks.
And this evolution also brings two new challenges to face recognition researches.
First, handheld devices have limited computing power and memory resources, which requires lightweight face recognition algorithms.
Second, mobile images and social network images are usually taken under uncontrolled imaging conditions,
which requires face recognition algorithms robust to a wide range of face variations.

Numerous algorithms have been proposed for face recognition. There are many different ways to categorize algorithms in face recognition.
Among them, two taxonomies are widely used: that is from feature representation perspective and from machine learning perspective.
The feature representation perspective considers how to represent faces with good features, and believes better features lead to better accuracy. A lot of features were invented for face recognition under this philosophy.
This includes handcrafted feature like Gabor \cite{EBGM,LiuGaborFRGC}, LBP\cite{LBPFace,LTPFRGC}, etc.
New features are still under emerging for face recognition.

The machine-learning perspective considers how to learn good representation/model for face recognition.
Methods in this taxonomy can be further divided into three categories.
First is feature learning, which tries to learn discriminant features from raw input.
This includes (1) subspace based methods \cite{EigenFace,Fisherface,LaplaceFace},
(2) mid-level representations \cite{FaceAttribute, SparseFace}, (3) deep learning based methods \cite{LEFace, DeepFace12}.
Feature learning is data-driven, that is the biggest difference to handcraft features.
Second, machine learning is used to learn a classification engine to separate faces from same subjects and different subjects.
The classification engine can be matching functions, distance metrics \cite{LDML,KISS,CSML},
and classifiers like SVM \cite{SVMFace} and Boosting \cite{BoostingFace}.
Third, machine learning is used on multiple results fusion, context learning, etc \cite{MLPQFRGC, OneShot}.

In this paper, we consider both the feature learning and classification engine learning within a consistent hierarchical framework.
Different from the unsupervised way in learning descriptor \cite{LEFace} or deep feature learning \cite{DeepFace12}, all the learning in the proposed approach are supervised.
Furthermore, the supervised learning was carried on a large-scale dataset to ensure robustness.
Major contributions of the proposed approach are as follows
\begin{itemize}
 \vspace{-0.05in}
 \small
 \addtolength{\itemsep}{-0.05in}
  \item[(1)] We develop a supervised hierarchical feature learning framework for face recognition, and demonstrate state-of-the-art performance on both the FRGC benchmark \cite{FRGC} and the LFW benchmark \cite{LFW}.
  \item[(2)] We do large-scale training on computing cluster, and show large-scale training really brings accuracy improvement.
  \item[(3)] We show that the proposed system has low computing cost and small memory footprint, which make it suitable for embedded devices.
 \vspace{-0.06in}
\end{itemize}

In the rest of the paper, we will first revisit related works in Section 2, and
present the framework overview in section 3. Details of hierarchical feature learning and large-scale training are presented in section 4 and 5.
Experiments are shown in Section 6. Conclusions are drawn in Section 7.

\section{Related Works}
We review related works in two aspects of face recognition: feature learning and classification engine learning.

\subsection{Feature Learning}
Feature learning in face recognition can be categorized into three major categories.
First, subspace methods try to cast the raw feature to a discriminant subspace, which
are dominant methods in face recognition researches in the past two decades.
Typical algorithms include eigen-faces\cite{EigenFace}, Fisherfaces \cite{Fisherface}, Lapacianfaces \cite{LaplaceFace},
and kernel subspace methods \cite{FaceKernel}. Subspace methods suffer from large projection matrix.
Suppose subspace methods project $d$-dimensional raw feature to $p$-dimensional discriminant subspace, the projection matrix is of size $d \times p$.
The raw feature dimension $d$ is usually very high. For instance, the dimension of Gabor features may be as high as tens of thousands \cite{LiuGaborFRGC,LTPFRGC}.
As a result, large projection matrix yields not only large memory footprint, but also high computing cost.
Some researches utilize the divided and conquer strategy, which divides the full feature vector into several blocks,
and solve projection in each block \cite{HECGaborFRGC, NECMetric}.
This paper applies subspace analysis to each patch descriptor, which can be viewed as a special case of block subspace analysis.
However, the computing complexity and memory footprint of patch-level subspace analysis is usually one order less
than that of block subspace methods.

Second is the methods that learn a mid-level representation in an unsupervised way,
which include learning descriptor \cite{LEFace}, deep feature learning\cite{DeepFace12} and sparse representation \cite{SparseFace}, etc,
The unsupervised learning is able to find common pattern from big data.
The proposed approach borrows the hierarchical architecture from these methods, but learns each feature layer in a supervised way.

Third, methods like attribute based algorithms \cite{FaceAttribute, TomPete} learn a mid-level representation in a supervised way.
However, they require an additional annotation of attributes or identity.
To guarantee scalability, the number of attributes should be large enough, and the number of annotated samples for each attribute should be sufficient.
This paper adopts supervised learning to learn hierarchical features with information just coming from face pairs.
More precisely, we only need information about whether two faces in given pair are from the same subject or not.
Besides, the derived feature of the proposed method does not have explicitly semantic meaning as that in attributes.

\subsection{Classification Engine Learning}
After the feature is obtained, one decision function is required to determine whether one input face belongs to a certain subject,
or whether input face pair (i.e., two faces) belongs to the same subject or not.
people usually use nearest neighbor classifier as the decision function.
In the early stage, the nearest neighbor decision was based on existing distance/similarities like Euclidean distance,
cosine similarity, etc. With the increase of training samples, learning based classification engines were emerged.
There are basically two categories.
First, the classification engine is built on each subject to classify whether the input face belongs to the subject or not.
 Classifiers like SVM, Boosting \cite{SVMFace, BoostingFace} were explored.
 This case requires each subject contains sufficient samples for training.
 The assumption is generally not true application like face verification.
\begin{figure*}[htbp]
 \vskip -0.05in
\footnotesize
 \centering
    \includegraphics[width=6.1in]{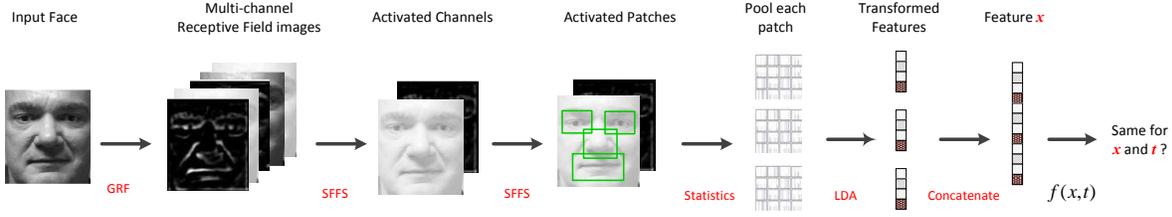}
 \caption{Flowchart of hierarchical feature learning for face recognition}
 \label{fig1}
 \vspace{-0.1in}
\end{figure*}

Second, metric learning tries to learn a metric which can maximally separate face pairs from same subjects and different subjects.
Given features of two faces $\mathbf{v}_i$ and $\mathbf{v}_j$,
the metric is usually in quadric form like
$d(\mathbf{v}_i, \mathbf{v}_j) = (\mathbf{v}_i - \mathbf{v}_j)^T \mathbf{M} (\mathbf{v}_i - \mathbf{v}_j)$,
where $\mathbf{M} \in R^{d\times d}$ is a symmetric positive defined matrix which can be decomposed as $\mathbf{M} = \mathbf{A}^T \mathbf{A}$.
The distance $d(\mathbf{v}_i,\mathbf{v}_j)$ can be embedded into objective functions like logistic discriminant function \cite{LDML}, cosine function \cite{CSML}.
 Optimal $\mathbf{M}$ or $\mathbf{A}$ is gotten by learning on the objective function,
 subject to pair-wise constraints \cite{LDML,CSML,KISS,NECMetric}.
 However, quadric metric has similar computing cost and memory cost as subspace based methods.

Recently, classification engine was applied to pair-wise features \cite{HTFFace,FaceAttribute}.
For features of two faces $\mathbf{v}_i$ and $\mathbf{v}_j$, we denote $\mathbf{x} \Leftarrow f(\mathbf{v}_i, \mathbf{v}_j)$ as the pair representation,
where $f(,)$ is an aggregation function, and should satisfy the symmetric property: $f(\mathbf{v}_i, \mathbf{v}_j) = f(\mathbf{v}_j, \mathbf{v}_i)$.
As in \cite{HTFFace,FaceAttribute}, function $f(,)$ can be element-wise absolute-difference
$\mathbf{x} = \mathbf{v}_i\ominus \mathbf{v}_j \equiv (|\mathbf{v}_{i1}- \mathbf{v}_{j1}|^p, \cdots, |\mathbf{v}_{id}- \mathbf{v}_{jd}|^p)$,
and element-wise product $\mathbf{x} = \mathbf{v}_i\odot \mathbf{v}_j \equiv (|\mathbf{v}_{i1}\cdot \mathbf{v}_{j1}|^p, \cdots, |\mathbf{v}_{id}\cdot \mathbf{v}_{jd}|^p)$.
Label $y$ for pair representation $\mathbf{x}$ is defined as: if $\mathbf{v}_i$ and $\mathbf{v}_j$ are from the same subject, y =1; otherwise y = -1.
Traditional classifier training can be directly applied to the pair-set $\{(\mathbf{x}, y)\}$.
Suppose there are $N$ faces in the original dataset $\{\mathbf{v}_i \}_{i=1}^N $, the number of pairs is $K$ = $N(N-1)/2$.
$K$ should be very large, which makes the pair-set even unable to fit into the memory of any machine, and thus makes the classifier training very difficult.

This paper adopts the pair-wise classification engine, and try to solve large-scale training problem in cluster.
\section{Overview of the Proposed Framework}
We aim at building a practical face recognition algorithm with small memory footprint and low computing cost.
Figure \ref{fig1} illustrates the framework of the proposed algorithm.
The system can be divided into two parts: hierarchical feature learning and large-scale classification engine training.

The hierarchical feature learning part consists of five sequential steps.
Suppose the input faces are detected, aligned according to facial landmarks and normalized to a standard size
(128$\times$128 pixels in our case).
\begin{itemize}
 \vspace{-0.05in}
 \small
 \addtolength{\itemsep}{-0.05in}
 \item[(1)] First, input face is modeled by receptive field theory, and represented by a multi-channel image,
      in which each channel is a Gaussian receptive response map at certain parameters.
 \item[(2)] Second, as some channels are more discriminant than others, we only activate top-$P$ most discriminant channels
      for face recognition via the floating search method.
 \item[(3)] Third, multi-channel face images are represented by over-completed local patches. As different patches have different discriminant power, we adopted floating search algorithm to choose top-$Q$ most discriminant local patches to activate.
 \item[(4)] Fourth, each activated patch is pooling over certain spatial cell structure to obtain feature vectors.
 \item[(5)] Fifth, feature descriptor of each patch is projected to lower dimension subspace with discriminant subspace analysis.
 \vspace{-0.05in}
\end{itemize}

Learned features of different patches are concatenated to obtain a full representation of the input face.
We extracted features for all the $N$ training face samples $\{\mathbf{v}_i\}_{i=1}^N$.
The large-scale classification engine is trained on pair-sets $\{(\mathbf{x},y)\}$.
As the size of pair-set is very large, we implemented the training algorithm on computing cluster.

\section{Hierarchical Feature Learning}
\subsection{Gaussian Receptive Maps}
We model face images with the receptive field theory.
Receptive field is a term in neuroscience, which is identified as the region of the visual cortex where light alters its firing.
Some vision researches show that receptive fields in the visual cortex can be well modeled using Gaussian derivative operators up to 4-th orders \cite{young2001gaussian}.
Following \cite{AgeRF,RFHist}, we refer to the Gaussian derivatives of images as the Gaussian receptive field (GRF) maps.
Given Gaussian function $G^{\sigma}(x,y)$ = $\exp\{-(x^2+y^2)/2\sigma^2\}$, the Gaussian derivatives are defined as
\begin{equation*}
\footnotesize
\label{eq0}
    G_{m,n}^{\sigma}(x,y) = \frac{\partial ^{m+n}}{\partial x^m \partial y^n} G^{\sigma}(x,y).
\end{equation*}
The Gaussian receptive map of image $I$ is defined as $L_{m,n}^{\sigma}(x,y) = G_{m,n}^{\sigma}(x,y) \otimes I(x,y)$,
where $\otimes$ denotes convolution operation, $m$ and $n$ are derivative order for horizontal and vertical directions.
The computing order of convolution and derivative operation can be exchanged, i.e.,
\begin{equation}
\footnotesize
\label{eq1}
    L_{m,n}^{\sigma}(x,y) = \frac{\partial ^{m+n}}{\partial x^m \partial y^n} \big(G^{\sigma}(x,y) \otimes I(x,y)\big).
\end{equation}

Each $L_{m,n}^{\sigma}$ can be viewed as one channel of a Gaussian receptive field image $I_{GRF}$ = $\{L_{m,n}^{\sigma}\}$.
According to the theory in \cite{AgeRF,RFHist}, $0< m+n \leq 4$, there are 14 different combinations of $m$ and $n$.
When defining smooth kernel size  $\{0, 3, 5, 7\}$ (here 0 means no smooth),
the number of channels is 56 (= 14 $\times$ 4).
We further allow diagonal and anti-diagonal gradients, the total number of channels reaches 112 (= 56 $\times$ 2).
It is obvious that not all channels are equal importance in face recognition.
Hence, it is necessary to activate only those most discriminant channels.

\subsection{Channel Activation}
To pick out most discriminant channels, we have to define features to describe each channel.
Here, we follow the scheme in \cite{LearningT2} to adopt S1 pooling and T2 transform.
The feature are obtained by pooling over 4$\times$4 spatial grids.
Each cell is represented by 2 values based on T2 transform \cite{LearningT2}: $\sum(|L_{m,n}^{\sigma}|+L_{m,n}^{\sigma})$ and
 $\sum (|L_{m,n}^{\sigma}|-L_{m,n}^{\sigma})$. This yields a feature vector of 32 dimension.
We further do the spatial pooling on each grid by sub-dividing it into 4$\times$4 sub-cells.
The two-layered pooling yields a feature vector of 544 (=32$\times$(1+16)) dimension.
We denote $f_{m,n}^{\sigma}$ as the 544-dimensional meta feature of the receptive map $L_{m,n}^{\sigma}$.
There are totally 112 such meta features.

The channel activation adopts sequential forward floating search (SFFS)\cite{SFFS} algorithm on these meta features $\{f_{m,n}^{\sigma}\}$.
The optimization objective is TPR (true-positive-rate) at FPR (false-positive-rate)=0.1\% with nearest neighbor classification.
Algorithm details are showed in Table \ref{tab1}.
\begin{table}[htbp]
  \footnotesize
  \vspace{-0.05in}
  \caption{SFFS for channels/patches activation}\label{tab1}
  \begin{tabular*}{3.25in}{p{3.20in}}
  \hline
    \begin{itemize}
    \vspace{-0.05in}
    \addtolength{\itemsep}{-0.03in}
    \item \textbf{Input}: channel-set/patch-set with corresponding feature set $F$ = $\{f_{m,n}^{\sigma}\}$.
        $J({F_k})$ to measure nearest neighbor classification accuracy based on feature ${F_k}$.
    \item \textbf{Initialize}: $F_0$ = $\emptyset$, $k$=0.
    \item \textbf{Step 1}: inclusion
        \begin{itemize}
        \addtolength{\itemsep}{-0.02in}
        \item{} Find best feature $f^{+}$ = $arg \max\limits_{f\in F\backslash F_k} J(F_k \cup f)$,
                where $F\backslash F_k$ means that $F$ excludes the subset $F_k$;
        \item{} $F_{k+1} = F_k \cup f^{+}$; $k$ = $k + 1$;
        \end{itemize}
    \item \textbf{Step 2}: conditional exclusion
        \begin{itemize}
        \addtolength{\itemsep}{-0.02in}
            \item{} Find worst feature $f^{-}$ = $arg \max\limits_{f\in F_k} J(F_k - f)$;
            \item{} if $J(F_k - f^{-})$ $>$ $J(F_{k-1})$
                \begin{enumerate}
                \addtolength{\itemsep}{-0.02in}
                  \item[-] $F_{k-1} = F_k - f^{-}$; $k$ = $k- 1$;
                  \item[-] goto \textbf{Step 2};
                \end{enumerate}
            \item{} else
                \begin{enumerate}
                \vspace{-0.02in}
                  \item[-] goto \textbf{Step 1};
                \vspace{-0.02in}
                \end{enumerate}
        \end{itemize}
     \item \textbf{Output}: channel/patch-subset corresponding to $F_k$.
     \vspace{-0.05in}
    \end{itemize} \\
    \hline
  \end{tabular*}
  \vspace{-0.1in}
\end{table}

\subsection{Patch Activation}
We further consider refining the location of receptive filed.
To realize this goal, we define over-complete patches based on picked channels following the strategy in \cite{NECMetric,JiaRF}.
For a 128$\times$128 face image and corresponding picked receptive maps,
we define a sliding window over it, and allow the window sliding 4 pixels forward .
The aspect-ratio of the sliding window can be 1:1, 1:2, 1:3, 1:4, 2:1, 3:1, 4:1, 2:3, 3:2.
We also adopt the 4$\times$4 spatial pooling for each patch, and restricted that each cell should contain at least 30 pixels.
Finally, we get a pool of about 10,000 patches.

Each patch is represented by spatial pooling features over multi-channel.
The feature dimension is 32$\times$$P$, where $P$ is the number of activated channels
 \footnote{Here 32 = 4$\times$4 spatial cell $\times$ 2 due to T2 transform.}.
With this patch descriptor, we followed the same scheme as that of channel activation to find most discriminant patches from the pool.

The reason of two-stage activation is two folds.
First, it is biologically motivated.
According to \cite{HMAX}, the brain uses a hierarchical approach for object recognition from simple layer to complex layer.
Second, the two-stage strategy is straightforward in computing.
If we consider finding best channels for each patch,
we had to face not only extremely large search space, but also additional computing cost due to
there would be no sharing computing of Gaussian receptive maps among different patches.

\subsection{Feature Pooling}
By default, we use the average pooling for each cell in one patch.
Generally, pooling is defined as accumulation of statistics for a set of samples (pixels).
This paper evaluates four different statistics in pooling stage.
\begin{itemize}
 \vskip -0.1in
 \small
 \addtolength{\itemsep}{-0.05in}
 \item[(1)]$\max$-pooling: it computes the maximum value in each cell $C_i$ of a patch, i.e.,
    $\max\limits_{(x,y) \in C_i}L_{m,n}^{\sigma}(x,y)$.
 \item[(2)]$\mu$-pooling (or average pooling): it computes the average value in each cell of a patch, i.e.,
    $\mu = E[L_{m,n}^{\sigma}(x,y)]$, where $E[x]$ means the expectation of variable $x$.
 \item[(3)]$\sigma$-pooling: it computes the variance value in each cell of a patch, i.e.,
    $\sigma^2 = E[(L_{m,n}^{\sigma}(x,y) - \mu)^2]$.
 \item[(4)]$m$-pooling: it computes the image moment value in each cell of a patch, i.e.,
    $\sum_{(x,y)\in C_i} {(x-x_c)^p (y-y_c)^q L_{m,n}^{\sigma}(x,y)}$,
    where $(x_c, y_c)$ is the center of the cell, $p$ and $q$ are the order over $x$ and $y$.
    And in this paper, we choose $p$=1 and $q$=1.
 \vspace{-0.08in}
\end{itemize}

Whatever pooling is adopted, the feature descriptor for each patch should be normalized.
We adopted SIFT-like normalization ($L_2$ normalization followed by clipping and renormalization),
and found it works the best.

\subsection{Discriminant Descriptors}
We have activated most discriminate channels, and most discriminate patches.
Furthermore, we can consider the discriminate capability within each patch.
We adopt linear discriminant analysis (LDA) to do patch-level subspace analysis.
We define pairwise intra-subject covariance matrix $S_w$ and extra-subject covariance matrix $S_b$ as
\begin{eqnarray}
 \footnotesize
 S_w = \sum\nolimits_{y_{ij}=1} (\mathbf{v}_i -\mathbf{v}_j)(\mathbf{v}_i -\mathbf{v}_j)^T, \nonumber \\
 S_b = \sum\nolimits_{y_{ij}=-1} (\mathbf{v}_i -\mathbf{v}_j)(\mathbf{v}_i - \mathbf{v}_j)^T, \nonumber
\end{eqnarray}
where $y_{ij}=1$ means that $\mathbf{v}_i$ and $\mathbf{v}_j$ comes from the same subject, otherwise $y_{ij}$ = $-1$.
The optimization objective is defined as  $J(\mathbf{w}) = { \mathbf{w}^T S_b \mathbf{w}}/{\mathbf{w}^T S_w \mathbf{w}}$.
There are many ways to solve this optimization problem.
In this paper, we adopt the enhanced fisher method by \cite{LiuGaborFRGC}.
In future, we may consider some maximum margin projection methods.

LDA will obtain a projection matrix $\mathbf{P}$ $\in$ $R^{d\times p }$, which projects the $d$-dimensional patch descriptor into $p$-dimensional discriminant subspace. The projected dimension $p$ is determined by the eigenvalue energy.
We keep the first $p$ dimension which the corresponding eigenvalues of LDA keep 99\% of total energy \cite{EigenFace}.
The projection matrix is learned from training set for each patch, and different patches can have different projected dimension $p$.

Learned patch descriptors from different patches are concatenated to obtain full representation of input faces.

\subsection{Computing/Memory Complexity}
The computing complexity of feature extraction lies in four major parts.
First, the computing complexity of Gaussian receptive maps is $O(P\cdot w \cdot h)$,
where $P$ is the number of activated channels, $w$ and $h$ are the size of normalized face image.
Second, the computing complexity of feature pooling is bounded by $O(P\cdot Q\cdot w \cdot h)$, while $Q$ is the number of activated patches.
To avoid redundant computing among different patches, integral image tricks can be used here.
Third, the computing complexity of patch feature projection is $O(Q\cdot d\cdot p)$,
where $d$ and $p$ are the original dimension and projected dimension of patch descriptor, respectively.
Fourth, the computing complexity of feature normalization is $O(Q\cdot d)$.
For a 128$\times$128 face image, suppose $P$ = 4, $Q$ = 240, $d$ = 128, $p$ = 100,
we find the overall feature extraction procedure requires about 5 MFlops.
This computing cost is affordable by embedded devices like smart phone.

The patch project matrix will cost $Q\cdot d \cdot p$ memory when quantization is enabled.
This is about 3MB when $Q$ = 240. And the size of linear SVM model is neglectable in comparison to this size.
Therefore, the overall memory footprint of the proposed method is very small.

\section{Learning Classification Engine}
\subsection{Pairwise Classification Engine}
We use pair-wise classification engine for face recognition.
Given face pairs $\mathbf{v}_i$ and $\mathbf{v}_j$, $\mathbf{x}$ is the pair representation which
can be element-wise absolute-difference or element-wise product as defined in section 2.2.
In practice, we find that $\mathbf{x} = \mathbf{v}_i\ominus \mathbf{v}_j$ works the best when setting $p= 0.5$.

The training set is reformulated as $\{(\mathbf{x},y)\}$ as described in section 2.2.
This paper adopts linear support vector machines (SVM) to train a classification engine over $\{(\mathbf{x},y)\}$.
The optimization goal for linear SVM is
\begin{equation}
 \footnotesize
\label{eq2}
    \frac{1}{2}\mathbf{w}^T \mathbf{w} + C\sum\nolimits_{i} \max(1-y_i\mathbf{w}^T \mathbf{x}_i, 0)^2,
\end{equation}
where $\mathbf{w}$ is the weights of linear SVM, and $C$ is tunable parameter for the regularization.

The objective can be optimized by many different methods as in \cite{LibLinear}.
With the learned classification engine, the recognition decision is still based on the nearest neighbor rule.
For a input face $\mathbf{v}$ and a template $\mathbf{t}$, the similarity is defined as $\mathbf{w}^T (\mathbf{v}\ominus \mathbf{t})$.

\subsection{Large-scale Training using ADMM}
Although there are many optimization methods for linear SVM, the training is still very difficult due to the scale.
Given $N$ faces in the training set, the number of face pairs are as many as $K$ = $N(N-1)/2$.
For instance in FRGC-204 dataset, the number of training samples is more than 37 millions, and the concatenated feature dimension is about 20,000. The whole dataset is thus more than 22TB ($K$ = 3.7e+8, $d$ = 20,000, $K$$\times$ $d$ $\times$4 = 22TB, here 4 is due to floating point precision) in storage. This is beyond the memory capacity of any single machine available today.

There are two different ways to handle this problem. First, we may do sampling or filtering to get a subset for training.
For a typical well-equipped workstation with 16GB RAM, it can handle about 200,000 samples, which is a very small portion (~5\textperthousand) of the whole pair-set. Hence, sampling can't catch all the variations in the whole training set. The model gotten in that way is far from optimum.
People also consider removing near duplicated samples from pairwise representation $\mathbf{x}$ with a filter.
However, the complexity of near duplicated filtering is $O(K^2)$, which is almost intractable on a single machine.
Besides, our experiments show that the portion of near duplicated samples is relative small (less than 20\% in FRGC-204).
Therefore, filtering will not change the scale of the training problem. Hence, sampling or filtering is not feasible.
In experiments, we will show sub-optimal result by the sampling method.

Second, we may try to use a computing cluster environment to employ large-scale training algorithms for this problem.
The progress of large-scale training has been emerged due to big data \cite{BigLearning}.
Among those emerging algorithms, stochastic gradient decent (SGD) and alternative direction method of multipliers (ADMM) are suitable for our task.
In practice, we choose ADMM since that it sufficiently utilizes each computing node, and converge much faster than SGD.

ADMM is applied to solve problems like: $\min f(\mathbf{w})+g(\mathbf{w})$.
 When $f(\mathbf{w})$ and $g(\mathbf{w})$ are of separate objective, and are difficult to optimize together
 due to function or data complexity, ADMM introduces a dual variable $\mathbf{z}$,
 and defines an equivalent constraint optimization problem as: $\min f(\mathbf{w}) + g(\mathbf{z}), ~s.t.~\mathbf{w}=\mathbf{z}$.
Although this change may seem trivial, the problem can now be attacked by the augmented Lagrangian methods.
In a nutshell, ADMM allows this problem to be solved approximately by first solving for $\mathbf{w}$ with $\mathbf{z}$ fixed,
and then solving for $\mathbf{z}$ with $\mathbf{w}$ fixed \cite{ADMM}, and repeating this dual updating procedure until convergency.

For the training problem in Eq \ref{eq2}, we divide the pair-set $\{(\mathbf{x},y)\}$ into $m$ blocks $\{B_1,\cdots,B_m\}$,
and distribute these blocks to different cluster nodes.
Under ADMM framework, the training objective can be rewritten as
\begin{eqnarray}
 \vspace{-0.06in}
 \footnotesize
 \label{eqadmm}
 \min_{\mathbf{w}_1,\cdots,\mathbf{w}_m, \mathbf{z}} &&\frac{1}{2}\mathbf{z}^T\mathbf{z} +C\sum_{j=1}^m\sum_{i\in B_j}\max(1-y_i\mathbf{w}_j^T\mathbf{x}_i, 0)^2 \nonumber\\
        &&+ \rho \sum_{j=1}^m \|\mathbf{w}_j -\mathbf{z}\|^2, \nonumber\\
        && s.t. ~~ \mathbf{w}_j- \mathbf{z} =0, \forall j
 \vspace{-0.06in}
\end{eqnarray}
where $\rho$ is pre-defined step for the optimization on dual variable $\mathbf{z}$.
For more details on training, please refer to \cite{ADMMLinar}.

The optimization of $\mathbf{w}_1,\cdots,\mathbf{w}_m$ can be decomposed into $m$ independent problems.
The optimization of $\mathbf{z}$ does not involve the training samples at all.
Hence, the training samples can be locally accessed in computing nodes and the communication cost is kept fairly low.

With ADMM training algorithms, we try to push the number of training samples used to the capacity limit of cluster.
We will show in experiments how more training data is useful to improve the accuracy.

\section{Experiments}
We implemented all the algorithm framework in C/C++, and the classification engine training part is
further implemented with MPI to make it able to run on cluster environment.
We conducted the training on a cluster with 320 computing nodes, in which
each node is a 2.6GHz Intel Xeon CPU with 64GB RAM.
We evaluated the proposed approach on two famous benchmarks: FRGC and LFW.

\subsection{The FRGC Benchmark}
We first studied the proposed approach in face recognition grand challenges (FRGC) version 2 \cite{FRGC}.
The protocol of FRGC is for face verification, i.e., whether the given face pairs are coming from same subject or not.
 Face images in FRGC2 are divided into training set and testing set without overlapped subjects.
 Our study is focused on FRGC2 experiment 4 (in simple, FRGC-204), which contains 12,776 training faces, 16,028 target face images and 8,014 query faces.
 The target images were obtained under controlled conditions but the query images were captured in uncontrolled settings.
 The big variations between target and query pose a real challenge to any recognition methods.
 The training set of FRGC-204 consists of 166,835 positive pairs, and 37,078,940 negative pairs.
 The experiment will produce three ROC curves (RoC-I, II, and III), corresponding to face images captured at different time,
 in which ROC-III is the most challenging task, and has been used as de facto metric to compare quality of algorithms
 in face recognition.

\begin{figure*}[htbp]
 \footnotesize
 \vskip -0.1in
 \centering
   \subfigure[]{\label{fig2a}
    \includegraphics[height=1.721in, width=2.2in]{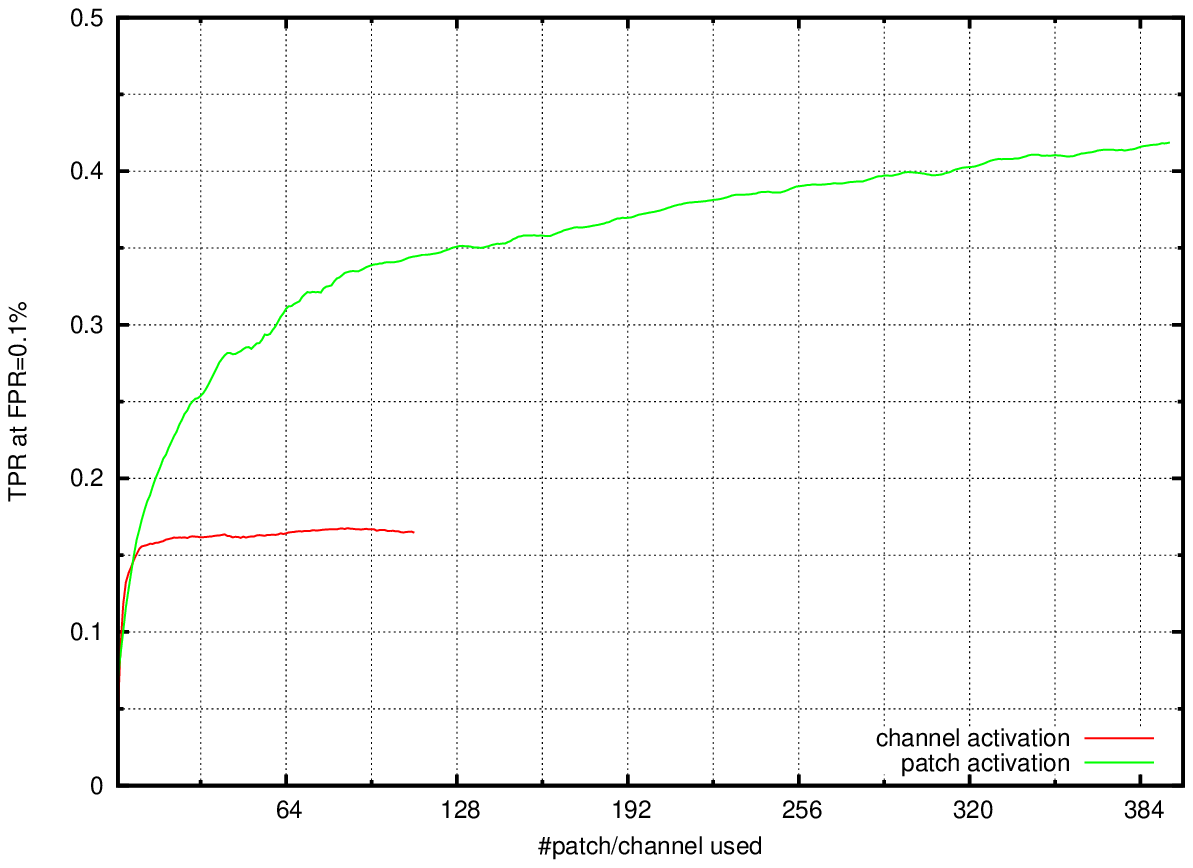}}
    \subfigure[]{\label{fig2b}
    \includegraphics[height=1.721in, width=2.2in]{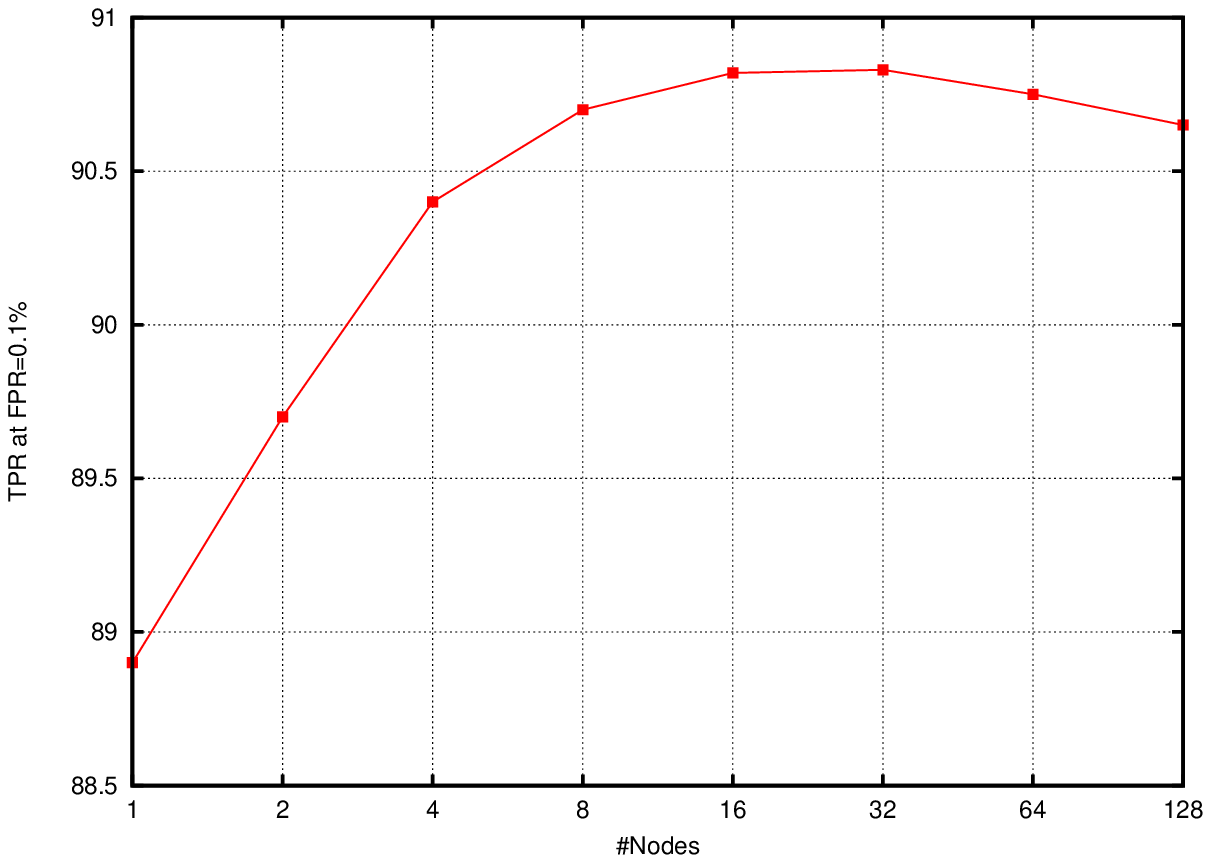}}
    \subfigure[]{\label{fig2c}
    \includegraphics[height=1.72in, width=2.3in]{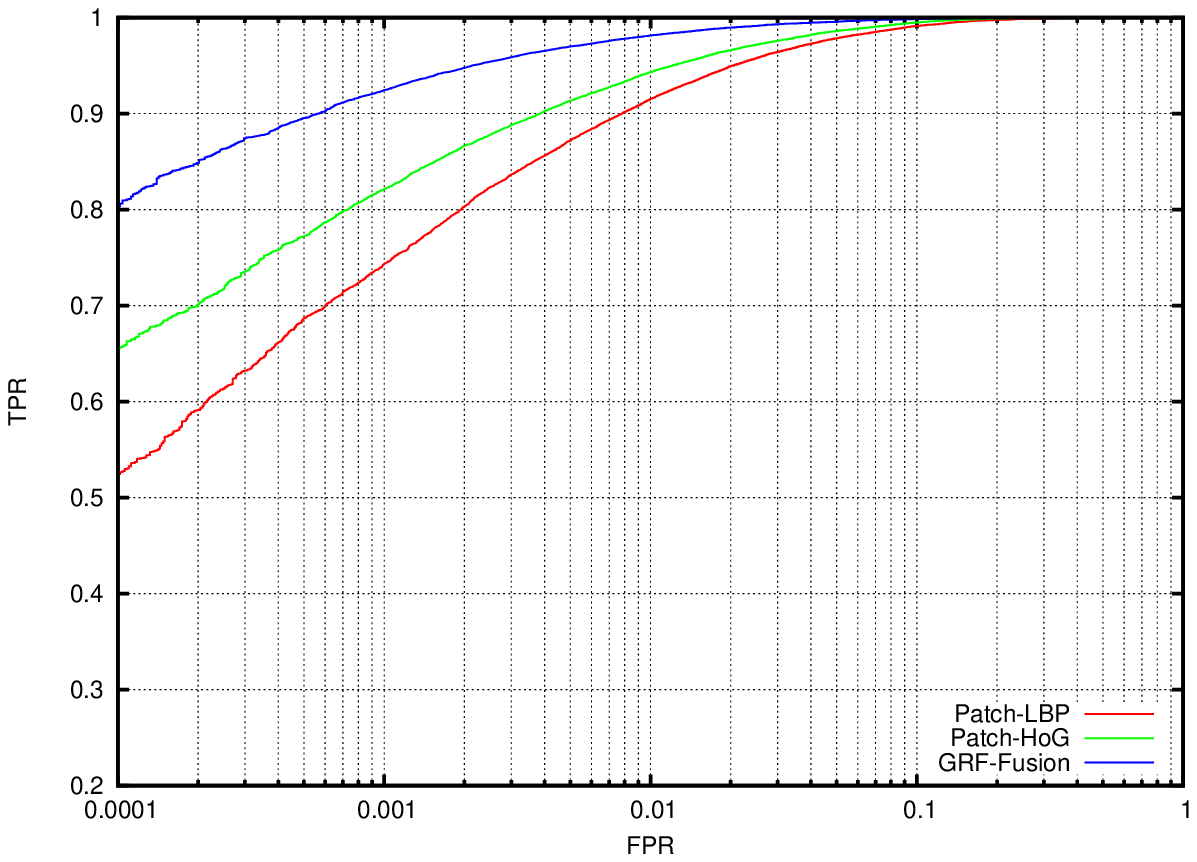}}
 \caption{(a) Red curve is TPR at FPR=0.1\% vs number of channels added by SFFS,
 while green curve is TPR at FPR=0.1\% vs number of patches added by SFFS;
 (b) TPR at FPR=0.1\% on FRGC-204 ROC-III with different computing nodes;
 (c) ROC-III for the proposed approach GRF-fusion on FRGC-204, in comparison to Patch-LBP and Patch-HoG.}
 \label{fig2}
  \vskip -0.1in
\end{figure*}

 We aligned and normalized faces with given landmarks in the dataset.
 We processed all the steps in Figure \ref{fig1}, and evaluated the proposed approach on ROC-III.
 We first divided the training set into two folds, and then used the SFFS algorithm to find the top-$P$
 most discriminant channels from a total of 112 Gaussian receptive maps with two-fold cross-validation.
 Figure \ref{fig2a} illustrates a curve with TPR at FPR=0.1\% vs number of channels on the testing set.
 In practice, we chose $P$ = 4 as it is a tradeoff between dimensionality and accuracy.

 We further used the SFFS algorithm to pick top-$Q$ most discriminant patches from an over-complete patch pool as described in section 4.3.
 Figure \ref{fig2a} illustrates another curve with TPR at FPR=0.1\% vs number of patches on the testing set.
 We chose $Q$=240 as it yields state-of-the-art accuracy in FRGC-204 with moderate computing complexity,
 though more patches may bring additional improvement.

 Given activated channels and patches, we extracted patch descriptor for all patches on the whole training set.
 By default, we use $\mu$-pooling for the patch description.
 Each patch is thus a 128-dimensional feature vector. We then trained an LDA projection for each patch from all training set
 according to the algorithm described in section 4.5.
 And descriptor of each patch is then projected to discriminate subspace.
 Samples of the same patch have the same projected dimension, but different patches may have different projected dimension.
 The average projected dimension is about 100 in practice. Hence, the concatenated feature vector for one face image is about 24,000 dimension.

 Large-scale training was performed on cluster.
 For negative pairs, we shuffle all pairs and divide them into equal-size subsets, and assign one subset to one node.
 As the number of positive pairs is much less than that of negative pairs,  we adopt bootstrap sampling to assign positive pairs for each node.
 The number of positive pairs for each node is 40,000, and the number of negative pairs for each node is 320,000
 \footnote{We have a tech-report/paper in parallel computing field, which illustrate how this configuration is selected by a lot of experiments.}.
 We distributed training subset locally at each node to avoid additional communication cost.
 The training can be converged with about 20 iterations on the dual updating of $\mathbf{z}$ in Eq \ref{eqadmm}.
 As the communication cost is very low, the training is dominant by model training in local node.

 Figure \ref{fig2b} illustrates the curve of TPR at FPR=0.1\% vs number of computing nodes used.
 Note that, we have made cost parameter ($C$ in SVM) tuning for each different number of nodes in this experiment.
 From this figure, we can have several conclusions.
 First, sampling based methods (result from one node) just produce a sub-optimal result.
 Second, more training samples really bring notable accuracy gains.
 Third, the accuracy will be saturated when the number of training samples reach a threshold.
 This is due to the fact that there are not only near-duplicated samples, but also samples with noise or non-linear variations.
 When most new samples are near-duplicated to existing ones, the training will not yield performance gain.
 When the portion of noise and non-linear variations samples are large enough, the accuracy may even degrade.
 In Figure \ref{fig2b}, the accuracy is saturated at 32 nodes, and starts degrading at 64 nodes.

 Besides the default $\mu$-pooling, we also tried the other three pooling methods listed in section 4.4.
 The training procedure follows the same scheme as that of $\mu$-pooling.
 Results on FRGC-204 benchmark are listed in Table \ref{tab2}.
 We can see that $\max$-pooling is the worst \footnote{This is due to the fact that
$\max$ value of a cell are very close among different faces.}, while $m$-pooling is the best here.
 We fused results by $\mu$-pooling, $\sigma$-pooling and $m$-pooling, to get a combined results
 \footnote{It is just a simple sum of the output scores by the classification engine of three pooling methods for each evaluated pair.}.
 This fusion further yields a notable accuracy improvement.  Figure \ref{fig2c} depicts the ROC-III curve by the fusion.

 Furthermore, we also made an experiment to show GRF is critical component in the hierarchical feature learning framework.
 We replace GRF with LBP and HoG in the proposed framework. One difference is that there are only patch activation for LBP and HoG.
 Hence, we called them Patch-LBP and Patch-HoG, respectively.
 In Patch-LBP, each patch is represented by a 59-dimensional uniform LBP histogram.
 And totally 400 patches are picked so that we make a fair comparison to GRF in terms of total feature dimension.
 In Patch-HoG, each patch is represented by a 128-dimensional feature (4$\times$4 cell $\times$ 8-dimensional histogram of oriented gradients in each cell). The results of Patch-LBP and Patch-HoG are also listed in Table \ref{tab2}.
 Besides, we listed some results by existing algorithms on the benchmark in Table \ref{tab2}.
 It is obvious that the proposed approach achieves a state-of-the-art result on FRGC-204.
 We should further pointed out that methods \cite{LiuGaborFRGC,LTPFRGC,HECGaborFRGC,MLPQFRGC}
 are all required extracting Gabor based features.
 Based on our evaluation, the proposed approach is an order of magnitude faster in feature extraction than Gabor based methods.
\begin{table}[htbp]
\vspace{-0.025in}
 \footnotesize
 \centering
 \caption{Result comparison on FRGC-204 (ROC-III) benchmark. Others' results are cited from corresponding papers.}
 \begin{tabular}{|c|c|c|c|}
  \hline
  Methods & TPR@FPR=0.1\%\\
  \hline
  \hline
  Baseline, eigenface \cite{FRGC} & 12\% \\
  \hline
  Gabor + Kernel \cite{LiuGaborFRGC} & 76\% \\
  \hline
  LTP + Gabor + Kernel \cite{LTPFRGC} & 88.5\% \\
 \hline
  Gabor + Fourier \cite{HECGaborFRGC} & 89\% \\
 \hline
   Method in \cite{MLPQFRGC}, single & 90.3\% \\
 \hline
   Method in \cite{MLPQFRGC}, multi-fusion & 91.6\% \\
  \hline
  \hline
  Patch-LBP & 74.3\% \\
 \hline
   Patch-HoG & 82.2\% \\
  \hline
  \hline
 GRF $\mu$-pooling & 90.7\% \\
  \hline
 GRF $\sigma$-pooling & 90.6\% \\
  \hline
   GRF $m$-pooling & 90.9\% \\
  \hline
    GRF $\max$-pooling & 77.3\% \\
  \hline
   GRF fusion($\mu$-,$\sigma$-,$m$-) & 92.6\% \\
  \hline
\end{tabular} \label{tab2} \vspace{-0.1in}
\end{table}

\subsection{The LFW Benchmark}
We further conducted experiments on the Labeled Faces in the Wild (LFW) benchmark \cite{LFW}, which consists of 13,233 images of 5,749 people.
The LFW benchmark is also for face verification, but is different from FRGC benchmark that
all images in LFW are collected from Internet.
\begin{table}[htbp]
\vspace{-0.025in}
 \footnotesize
 \centering
 \caption{Result comparison (average accuracy $\pm$ standard error)
    on LFW under \textit{image-restricted with label-free outside data} protocol.}
 \begin{tabular}{|c|c|c|c|}
  \hline
  Methods & Accuracy (\%)\\
  \hline
  \hline
  LDML \cite{LDML} & 79.27$\pm$0.60 \\
  \hline
  One-Shot \cite{OneShot} & 83.98$\pm$0.35 \\
  \hline
   CSML+SVM \cite{CSML} & 88.00$\pm$0.37 \\
 \hline
   Single LE \cite{LEFace} & 81.22$\pm$0.53 \\
 \hline
   High-Throughput \cite{HTFFace} & 88.13$\pm$0.58 \\
 \hline
   SFRD+PMML \cite{SFRD13} & 89.35$\pm$0.50 \\
  \hline
   VMRS\cite{VMRS} & 91.10$\pm$0.59 \\
  \hline
  \hline
 GRF $\mu$-pooling & 90.17$\pm$0.51 \\
  \hline
 GRF $\sigma$-pooling & 89.19$\pm$0.67 \\
  \hline
 GRF $m$-pooling & 89.93$\pm$0.58 \\
  \hline
 GRF fusion($\mu$-,$\sigma$-,$m$-) & 91.54$\pm$0.49 \\
  \hline
\end{tabular} \label{tab3}
\vspace{-0.1in}
\end{table}

\begin{figure}[tbp]
 \vskip -0.1in
 \footnotesize
 \centering
    \includegraphics[width=3.0in]{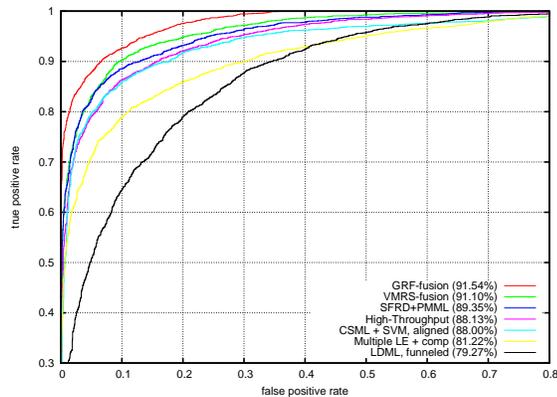}
    \caption{ROC curve of the proposed approach in comparison to existing algorithms on LFW under image restricted case.}
 \label{fig3}
  \vskip -0.1in
\end{figure}

In this benchmark, we used LFW-a (the aligned version of LFW), and cropped and resized faces to 150$\times$80
based on eye-centers according to suggestion in \cite{VMRS}.
We followed the same training framework of the FRGC experiments.
We conducted experiments strictly under the \textit{image-restricted with label-free outside data} protocol.
The protocol did 10-fold cross validation.
The whole dataset are divided into 10 subsets.
On the 9 training subsets, we did the channel selection and patch selection.
In this experiments, we selected top-4 channels and top-200 patches for all cross-validation trials.
We further trained the patch LDA projection and the SVM classification engine on 9 training subsets,
and tested the accuracy on the left one subset. Repeating this procedure 10 times,
we got the average accuracy and standard error as listed in Table \ref{tab3}, and ROC curve as illustrated in Figure \ref{fig3}.
The table and figure also include existing results obtained under the same evaluation protocol.
It is obvious that the proposed approach outperforms existing algorithms under the same protocol.

Note in this experiment\footnote{We did it based on the LFW updated protocol in April, 2014!}, the training pairs are fairly small in each cross-validation fold(5400). Training can be done even on a desktop. In the future, we will try the unrestricted with labeled outside data protocol,
and train with large-scale outside dataset to further increase the recognition accuracy on LFW.

\section{Conclusion}
   This paper proposes a novel face recognition algorithm based on large-scale supervised hierarchical feature learning.
   We first perform hierarchical feature learning from channel granularity to patch granularity, and further to descriptor granularity.
   We then train a linear classifier over pair representation to separate face pairs from same subject and different subject.
   As the number of face pairs are extremely large, we introduce ADMM to train the linear classifier on a computing cluster.
   Experiments show that more training samples will bring notable accuracy improvement.

   We conduct experiments on two famous benchmarks: FRGC and LFW.
   The proposed approach outperforms existing algorithms notably.
   Besides, the proposed approach is small in memory footprint, and low in computing cost, which makes it suitable for embedded applications.

{\footnotesize
\bibliographystyle{ieee}
\bibliography{deep13}
}

\end{document}